\documentclass[runningheads]{llncs}
\usepackage{graphicx}
\usepackage{multirow}
\usepackage{colortbl}
\newcommand{\tabincell}[2]{\begin{tabular}{@{}#1@{}}#2\end{tabular}}
\usepackage{array}
\usepackage{bm}
\usepackage{color}
\usepackage{amssymb}
\usepackage[colorlinks,
            linkcolor=red,
            anchorcolor=blue,
            citecolor=blue
            ]{hyperref}
\usepackage{bbm}
\usepackage{booktabs}
\usepackage{mathptmx}
\usepackage[font=footnotesize,labelfont=bf]{caption}
\usepackage{marvosym}

\begin{document}
\title{Multi-frame Collaboration for Effective Endoscopic Video Polyp Detection via Spatial-Temporal Feature Transformation}
\titlerunning{Spatial-Temporal Feature Transformation}
%
\author{Lingyun Wu\inst{1}
\and
Zhiqiang Hu\inst{1}
\and
Yuanfeng Ji\inst{2}
\and
Ping Luo\inst{2}
\and
Shaoting Zhang\inst{1}
\textsuperscript{(\Letter)}}
%
\authorrunning{L. Wu et al.}
%
%
\institute{SenseTime Research\\
\email{zhangshaoting@sensetime.com}\\
\and The University of Hong Kong\\
}
\maketitle              
\begin{abstract}
Precise localization of polyp is crucial for early cancer screening in gastrointestinal endoscopy.
Videos given by endoscopy bring both richer contextual information as well as more challenges than still images.
The camera-moving situation, instead of the common camera-fixed-object-moving one, \hyphenation{leads}leads to significant background variation between frames.
Severe internal artifacts (e.g. \hyphenation{water}water flow in the human body, specular reflection by tissues) can make the quality of adjacent frames vary considerately.
These factors hinder a video-based \hyphenation{model}model to effectively aggregate features from neighborhood frames and give better predictions.
In this paper, we present Spatial-Temporal Feature Transformation (STFT), a multi-frame collaborative framework to address these issues.
\hyphenation{Spatially}Spatially, STFT mitigates inter-frame variations in the camera-moving situation with feature alignment by proposal-guided deformable convolutions.
Temporally, STFT proposes a channel-aware attention module to simultaneously estimate the quality and correlation of adjacent frames for adaptive feature aggregation.
\hyphenation{Empirical}Empirical studies and superior results demonstrate the effectiveness and stability of our method.
For example, STFT improves the still image baseline FCOS by $10.6\%$ and $20.6\%$ on the comprehensive F1-score of the polyp localization task in CVC-Clinic and ASUMayo datasets, respectively, and outperforms the state-of-the-art video-based method by $3.6\%$ and $8.0\%$, respectively.
%
Code is available at \url{https://github.com/lingyunwu14/STFT}.
\end{abstract}

\section{Introduction}
Gastrointestinal endoscopy is widely used for early gastric and colorectal cancer screening, during which a flexible tube with a tiny camera is inserted and guided through the digestive tract to detect precancerous lesions \cite{ali2021deep}.
Identifying and removing adenomatous polyp are routine practice in reducing gastrointestinal cancer-based mortality \cite{jemal2008cancer}.
However, the miss rate of polyp is as high as $27\%$ due to subjective operation and endoscopist fatigue after long duty \cite{ahn2012miss}.
An automatic polyp detection framework is thus desired to aid in endoscopists and reduce the risk of misdiagnosis.

For accurate and robust polyp detection, it is necessary to explore the correlation and complementarity of adjacent frames, to compensate for the possible image corruption or model errors in single images \cite{qadir2019improving}.
Nevertheless, there have been two long-standing and serious challenges in endoscopic video polyp detection:

\begin{figure}[t]
\centering
\includegraphics[width=0.98\textwidth]{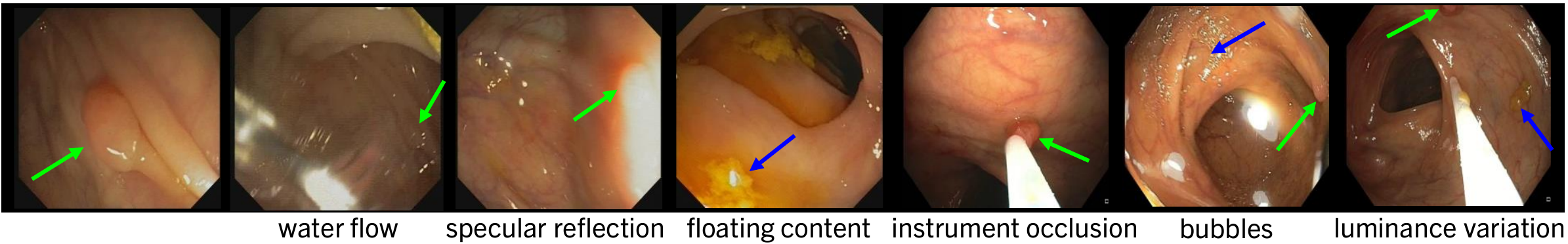}
\caption{The easier frame (the first) and typical challenging frames (last six) associated with polyp detection in endoscopic video.
Green arrows point to the polyp and blue arrows point to the internal artifacts that often cause false positives.} \label{fig_challenge}
\end{figure}

{\itshape How to align object features across frames given the complex motion of the endoscopic camera?}
One key difference of endoscopic videos from common videos is the camera-moving instead of the common camera-fixed-object-moving situation.
The complex motion of the endoscopic camera leads to significant background variation between frames.
As a result, the mainstream video methods \cite{zhu2017flow,zhu2017deep} based on optical flow alignment are not suitable, since you do not have a reference, which leads to poor performance of optical flow evaluation \cite{zheng2019polyp}.
Intuitively, attempting a global alignment is both difficult and unnecessary, since the variable background distracts the focus of the network and overwhelms the foreground modeling for feature alignments.
We thus argue that an object-centered and proposal-guided feature alignment is required to mask out background trifles and focus on the concerned foreground variation.

{\itshape How to assemble features of neighborhood frames given the varied image quality resulting from water flow, reflection, bubbles, etc.?}
As shown in Fig.~\ref{fig_challenge}, frames in endoscopic videos are always and inevitably encountered with image corruptions such as water flow, specular reflection, instrument occlusion, bubbles, etc.
These internal artifacts can make the quality of adjacent frames vary considerately.
The \emph{quality}, as a result, should be given equal consideration as the \emph{correlation} between frames, in the stage of adjacent feature aggregation.
We further notice that different internal artifacts are handled by different kernels in convolutional neural networks, and results in varied activation patterns in different channels.
The combination of channel-by-channel selection and position-wise similarity \cite{bertasius2018object,deng2019relation} are thus believed to be necessary for the simultaneous assessment of foreground correlation and feature quality.

We aim to tackle the two challenges with carefully designed spatial alignment and temporal aggregation, and propose the multi-frame collaborative framework named Spatial-Temporal Feature Transformation (STFT).
Spatially, we choose deformable convolution \cite{dai2017deformable} as building blocks for feature alignments, for its adaptability in modeling large variations.
We further enhance its object-centered awareness and avoid background distraction by conditioning the offset prediction of the deformable convolution on the object proposals extracted by the image-based detector.
Temporally, we design a channel-aware attention module that combines both the cosine similarity to model the foreground correlation between frames and the learned per-channel reweighting to estimate the inter-frame quality variation.
The modules achieve a balance of expressiveness and efficiency without much additional computational complexity.
Note that the two components are mutually beneficial in that spatial alignment acts as the prerequisite and temporal aggregation looks for more advantage, which is also demonstrated in experimental results.

The contribution of this work can be summarized as three folds.
Firstly, we present a proposal-guided spatial transformation to enhance the object-centered awareness of feature alignment and mitigate the feature inconsistency between adjacent frames in the camera-moving situation of endoscopic videos.
%
Secondly, we design a novel channel-aware attention module for feature aggregation that achieves a balance of expressiveness and efficiency and shows superiority over other counterparts in experimental results.
Lastly, we propose an effective multi-frame collaborative framework STFT on top of the two components.
STFT sets new state-of-the-arts on both two challenging endoscopic video datasets and two polyp tasks.
Noticeably, STFT shows a far more significant improvement over still image baselines than other video-based counterparts (for example, $10.6\%$ and $20.6\%$ localization F1-score improvements on the CVC-Clinic and ASUMayo datasets, respectively).

\begin{figure}[t]
\centering
\includegraphics[width=0.95\textwidth]{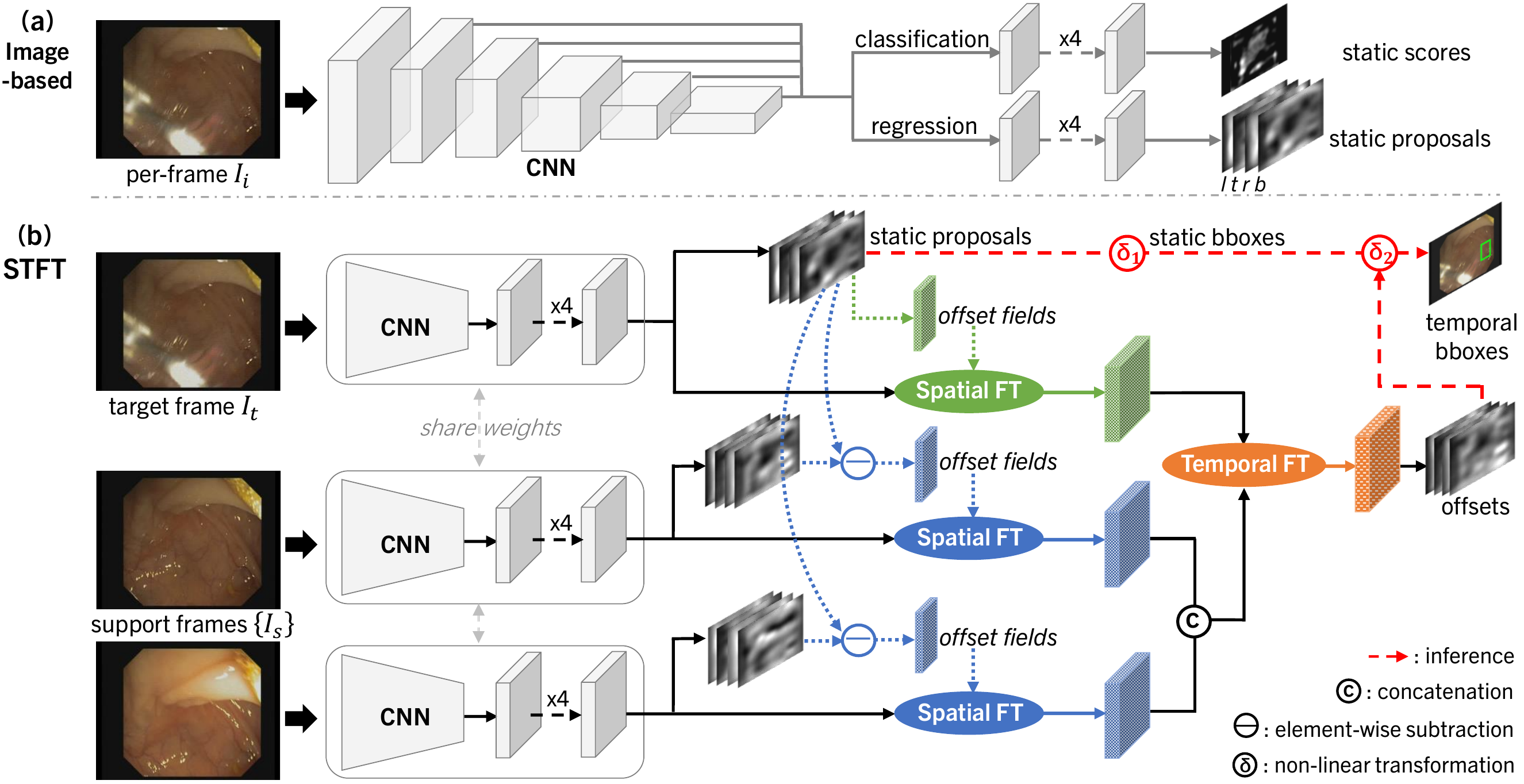}
\caption{Illustration of polyp detection using an image-based baseline (a) and our proposed Spatial-Temporal Feature Transformation (b). See section~\ref{method} for more details.} \label{fig_framework}
\end{figure}

\section{Method} \label{method}
\subsubsection{An Image-based Baseline}
Given the endoscopic video frames $\{I_i\}, i=1, ..., \infty$, a baseline approach for polyp detection is to apply an image-based detector to each frame individually.
We adopt a simple one-stage detector FCOS \cite{tian2019fcos} as our baseline.
As Fig.~\ref{fig_framework}(a), it firstly generates a set of multi-level feature maps with FPN \cite{lin2017feature} over the input image $I_i$.
Then, it outputs static classification scores and regression proposals by classification branch and regression branch respectively.
Each branch is implemented by four convolutional layers, sharing weights between different feature levels.
For level $l$, Let $F_i^l \in \mathbb{R}^{C \times H\times W}$ be the final feature map output by the regression branch.
$G_i=(x_0, y_0, x_1, y_1)$ is the associated ground-truth bounding box, where $(x_0, y_0)$ and $(x_1, y_1)$ denote the coordinates of the left-top and right-bottom corners.
Static proposals are $\{\bm{p}_i\} \in \mathbb{R}^{4 \times H\times W}$, and their regression targets $\{\bm{g}_i\} \in \mathbb{R}^{4 \times H\times W}$ are offsets between $G_i$ and all spatial locations on $F_i^l$.
For each spatial location $(x,y)$, $\bm{g}_i=(l_i, t_i, r_i, b_i)$ is a $4D$ real vector, which represents distances from $(x,y)$ to four boundaries of $G_i$.
It can be formulated as
\begin{equation} \label{eq1}
l_i = x - x_0,  t_i = y - y_0,  r_i = x_1 - x,  b_i = y_1 - y.
\end{equation}
Given the challenging frame with water flow in Fig.~\ref{fig_challenge}, the baseline showed low confidence on the ground-truth and missed polyp detection (see results in Fig.~\ref{fig_qa_image_visu}).

\subsubsection{STFT Architecture}
Given the target frame $I_t$ and its adjacent support frames $\{I_s\}, s=1, ..., N$, our aim is to accurately detect polyp in $I_t$ by using the features from $\{I_s\}$.
Firstly, we generate multi-level feature maps and static prediction results for all frames via the same architecture as the image-based baseline (only the regression branch is shown in Fig.~\ref{fig_framework}(b) for ease of explanation).
At $l$-th feature level, we use predicted static proposals $\{\bm{p}_t\}$ of $I_t$ to guide the spatial transformation of target feature $F_t^l$ (as shown in green in Fig.~\ref{fig_framework}(b)).
Meanwhile, we leverage the difference between $\{\bm{p}_t\}$ and static proposals $\{\bm{p}_s\}$ of each $I_s$ to guide the spatial transformation of each support feature $F_s^l$ to align it with the target (blue operations in Fig.~\ref{fig_framework}(b)).
Then, we model channel-aware relations of all spatially aligned features via a temporal feature transformation module (orange in Fig.~\ref{fig_framework}(b)).
For each level, the temporal transformed features of the classification branch and the regression branch predict offsets for static scores and static proposals, respectively.
The ultimate temporal bounding box are computed in non-linear transformations between static proposals and proposal offsets (red dashed line in Fig.~\ref{fig_framework}(b)), while the ultimate classification scores are obtained by multiplying the static scores and score offsets.
Finally, the predictions from all levels are combined using non-maximum suppression just like FCOS.

\subsubsection{Proposal-guided Spatial Feature Transformation} \label{shape_guided}
Ideally, the feature for a large proposal should encode the content over a large region, while those for small proposals should have smaller scopes accordingly \cite{wang2019region}.
Following this intuition, we spatially transform $F_t^l$ based on proposals $\{\bm{p}_t\}$ to make the feature sensitive to the object.
In practice, the range of each $\bm{p}_t$ is image-level.
In order to generate feature-level offset fields required for each spatial location deformation, we first calculate normalized proposals $\bm{p}_t^*=(l_t^*, t_t^*, r_t^*, b_t^*)$ with
\begin{equation}
l_t^* = \frac{-l_t}{s_l},  t_t^* = \frac{-t_t}{s_l},  r_t^* = \frac{r_t}{s_l},  b_t^* = \frac{b_t}{s_l},
\end{equation}
where $s_l$ is the FPN stride until the $l$-level layer.
We devise a $1\times 1$ convolutional layer $\mathcal N_{o}$ on $\{\bm{p}_t^*\}$ to generate the proposal-guided offset fields and a $3\times 3$ deformable convolutional layer $\mathcal N_{dcn}$ to implement spatial feature transformation, as follows:
\begin{equation}
F_{t}^{l\prime} = \mathcal N_{dcn}(F_t^l, \mathcal N_{o}(\{\bm{p}_t^*\})),
\end{equation}
where $F_{t}^{l\prime}$ is spatial transformed features.
We also perform this spatial transformation scheme on support features $\{F_s^l\}$.
Specially, we propagate predicted proposals from $I_t$ to each $I_s$ and leverage the difference between them to generate offset fields for deformation of each $F_s^l$.
In other words, we make each spatial transformed feature $F_s^{l\prime}$ sensitive to both the object and the difference.
This step plays a key role in improving the recall rate (see Table~\ref{tab_module_design}).

\begin{figure}[t]
\centering
\includegraphics[width=0.95\textwidth]{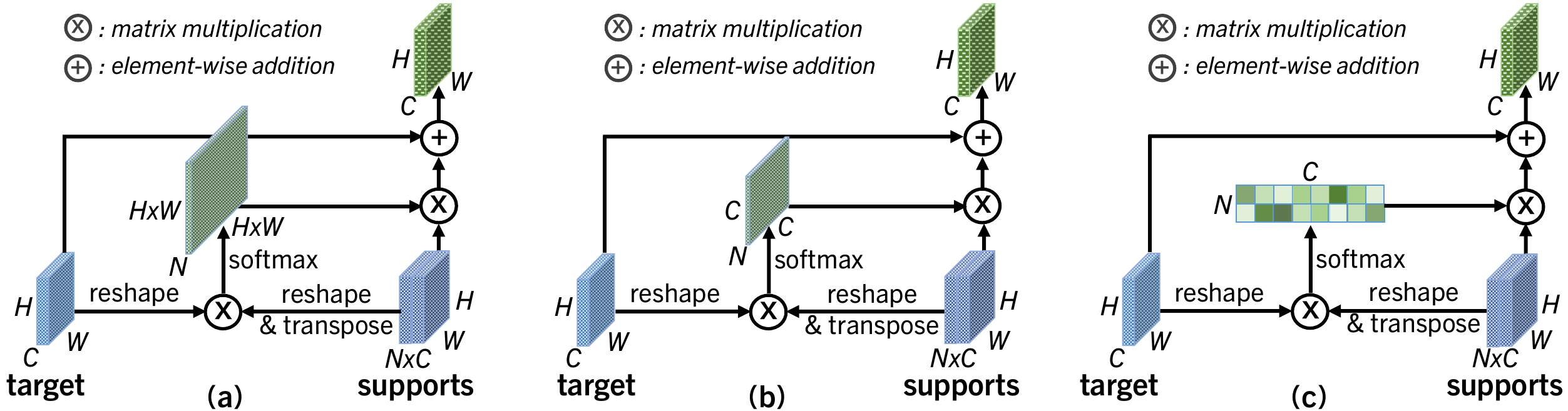}
\caption{Comparisons of different attention mechanisms. (a) Point-wise. (b) Channel-wise. (c) proposed Channel-aware for Temporal Feature Transformation.} \label{fig_attention}
\end{figure}

\subsubsection{Channel-aware Temporal Feature Transformation} \label{met_attention}
Because $I_t$ and $\{I_s\}$ share weights of all layers in our STFT, it can be considered that features of all frames on each channel have been spatially aligned after spatial feature transformation.
On the other hand, based on the principle of deep convolutional network learning, the features of certain channels are bound to be sparse, and their activations are close to zero.
Naturally, we propose channel-aware temporal feature transformation, aiming to mine the most representative channel features in the neighborhood for feature aggregation.
It is implemented by the attention mechanism.
We calculate the channel-aware attention map $A_{ts}^l$ from spatial aligned features $F_t^{l\prime}$ and $\{F_s^{l\prime}\}$ by
\begin{equation}
A_{ts}^l=softmax(\frac{\mathcal R(F_t^{l\prime}) \mathcal R(\{F_s^{l\prime}\})^T}{\sqrt {d_{f}}}) \mathcal R(\{F_s^{l\prime}\}),
\end{equation}
where $\mathcal R$ is the reshape and $T$ is the transpose for matrix multiplication, more details shown in Fig.~\ref{fig_attention}(c).
$d_{f}$ is a scaling factor \cite{vaswani2017attention}.
In our algorithm, $d_{f}$ is equal to $H\times W$ that represents the dimension of each channel feature.

\subsubsection{Target Assignment}
According to Eq.~\ref{eq1}, for each location $(x, y)$, we can obtain the predicted static bounding box $\bm{y}_t=(x-l_t, y-t_t, x+r_t, y+b_t)$ from static proposals $\bm{p}_t$.
In our STFT, if the intersection-over-union between $\bm{y}_t$ and $G_t$ is larger than a threshold ($0.3$ by default), the temporal classification label of $(x, y)$ is assigned to positive and $\bm{p}_t$ is considered a significative proposal guide.
Then, the temporal regression target $\bm{\delta}_t$ for $(x,y)$ is offsets between $\bm{y}_t$ and $G_t$.
$\bm{\delta}_t=({\delta}_{x_0}, {\delta}_{y_0}, {\delta}_{x_1}, {\delta}_{y_1})$ are computed by
\begin{footnotesize}
\begin{equation}
{\delta}_{x_0} = \frac{x_0-(x-l_t)}{w*\sigma},  {\delta}_{y_0} = \frac{y_0-(y-t_t)}{h*\sigma},  {\delta}_{x_1} = \frac{x_1-(x+r_t)}{w*\sigma},  {\delta}_{y_1} = \frac{y_1-(y+b_t)}{h*\sigma},
\end{equation}
\end{footnotesize}
where $w, h$ are the width and height of $\bm{y}_t$, and $\sigma = 0.5$ is the variance to improve the effectiveness of offsets learning.

\subsubsection{Loss Function}
Objects with different sizes are assigned to different feature levels.
Combining outputs from each level, our STFT is easy to optimize in an end-to-end way using a multi-task loss function as follows:
\begin{equation}
\mathcal L = \mathcal L_{cls} + \mathcal L_{reg} + \frac{1}{\mathcal N_{pos}} \sum_{x,y} \mathcal L_{cls}^{\bm{st}} (\mathcal C^{\bm{st}}, \mathcal C^*) + \mathcal L_{reg}^{\bm{st}} \mathbbm{1}_{\{\mathcal C^*>0 \} } (\Delta ^{\bm{st}}, \Delta ^*),
\end{equation}
where $\mathcal L_{cls}$ and $\mathcal L_{reg}$ are the static classification and regression loss respectively \cite{tian2019fcos}.
%
$\mathcal L_{cls}^{\bm{st}}$ is the temporal classification loss implemented by focal loss and $\mathcal L_{reg}^{\bm{st}}$ is the temporal regression loss implemented by $\mathcal L_1$ loss.
$\mathcal C^{\bm{st}}$ and $\Delta ^{\bm{st}}$ are predicted offsets for scores and proposals by STFT.
$\mathcal C^*$ and $\Delta ^*$ are assigned classification label and regression target.
$\mathbbm{1}_{\{\mathcal C^*>0 \} }$ is the indicator function, being 1 if $\mathcal C^*>0$ and 0 otherwise.

\begin{table}[t]
\caption{Quantitative comparison with SOTA Image-based and Video-based methods on CVC-Clinic and ASUMayo video datasets. The subscript list the relative gains compared to the corresponding Image-based baseline. `$N/A$' denotes not available.}\label{tab_polyp}
\centering
\resizebox{0.95\columnwidth}{!}{
\smallskip\begin{tabular}{p{0.7cm}<{\centering}p{1.1cm}<{\centering}p{0.5cm}<{\raggedright}p{4.4cm}<{\raggedleft}|p{1.5cm}<{\centering}p{1.5cm}<{\centering}p{1.6cm}<{\centering}|p{1.5cm}<{\centering}p{1.5cm}<{\centering}p{1.6cm}<{\centering}}
\toprule[2pt]
\multicolumn{4}{c|}{~} & \multicolumn{3}{c|}{Polyp Detection} & \multicolumn{3}{c}{Polyp Localization} \\
~ & ~ & $\#$ & Methods & Precision & Recall & F1-score & Precision & Recall & F1-score \\ \midrule[1.5pt]
\multirow{10}*{\rotatebox{90}{{\bfseries CVC-Clinic}}} & \multirow{5}*{\tabincell{c}{$Image$ \\ -$based$}} & $1$ & UNet~\cite{ronneberger2015u} $(MICCAI'15)$ & $89.7$ & $75.9$ & $82.2$ & $81.7$ & $72.0$ & $76.5$ \\
~ & ~ & $2$ & Faster R-CNN~\cite{girshick2015fast} $(ICCV'15)$ & $84.6$ & $98.2$ & $90.9$ & $78.5$ & $87.9$ & $82.9$ \\
~ & ~ & $3$ & R-FCN~\cite{dai2016r} $(NIPS'16)$ & $91.7$ & $87.1$ & $89.3$ & $81.4$ & $83.2$ & $82.3$ \\
~ & ~ & $4$ & RetinaNet~\cite{ross2017focal} $(CVPR'17)$ & $93.7$ & $86.2$ & $89.8$ & $87.8$ & $83.1$ & $85.4$ \\
~ & ~ & $5$ & Yolov3~\cite{redmon2018yolov3} $(arXiv'18)$ & $N/A$ & $N/A$ & $N/A$ & $98.3$ & $70.5$ & $82.1$ \\
~ & ~ & $6$ & FCOS~\cite{tian2019fcos} $(ICCV'19)$ & $92.1$ & $74.1$ & $82.1$ & $94.7$ & $70.4$ & $80.8$ \\
~ & ~ & $7$ & PraNet~\cite{fan2020pranet} $(MICCAI'20)$ & $94.8$ & $82.2$ & $88.1$ & $96.7$ & $82.1$ & $88.8$ \\ \cline{3-10}
~ & \multirow{5}*{\tabincell{c}{$Video$ \\ -$based$}} & $3^*$ & FGFA~\cite{zhu2017flow} $(ICCV'17)$ & $\bm{94.5}_{\uparrow2.8}$ & $89.2_{\uparrow2.1}$ & $91.7_{\uparrow2.4}$ & $88.7_{\uparrow7.3}$ & $86.4_{\uparrow3.2}$ & $87.6_{\uparrow5.3}$ \\
~ & ~ & $2^*$ & RDN~\cite{deng2019relation} $(ICCV'19)$ & $91.2_{\uparrow6.6}$ & $91.3_{\downarrow6.9}$ & $91.2_{\uparrow0.3}$ & $88.7_{\uparrow10.2}$ & $85.9_{\downarrow2.0}$ & $87.3_{\uparrow4.4}$ \\
~ & ~ & $1^*$ & OptCNN~\cite{zheng2019polyp} $(ISBI'19)$ & $84.6_{\downarrow5.1}$ & $\bm{97.3}_{\uparrow\bm{21.4}}$ & $90.5_{\uparrow8.3}$ & $74.3_{\downarrow7.4}$ & $\bm{96.4}_{\uparrow\bm{24.4}}$ & $83.9_{\uparrow7.4}$ \\
~ & ~ & $5^*$ & AIPDT~\cite{zhang2020asynchronous} $(MICCAI'20)$ & $N/A$ & $N/A$ & $N/A$ & $90.6_{\downarrow7.7}$ & $84.5_{\uparrow14.0}$ & $87.5_{\uparrow5.4}$ \\
~ & ~ & $2^{**}$ & MEGA~\cite{chen2020memory} $(CVPR'20)$ & $91.6_{\uparrow\bm{7.0}}$ & $87.7_{\downarrow10.5}$ & $89.6_{\downarrow1.3}$ & $91.8_{\uparrow\bm{13.3}}$ & $84.2_{\downarrow3.7}$ & $87.8_{\uparrow4.9}$ \\
~ & ~ & $6^*$ & {\bfseries STFT $(Ours)$} & $91.9_{\downarrow0.2}$ & $92.0_{\uparrow17.9}$ & $\bm{92.0}_{\uparrow\bm{9.9}}$ & $\bm{95.0}_{\uparrow0.3}$ & $88.0_{\uparrow17.6}$ & $\bm{91.4}_{\uparrow\bm{10.6}}$ \\ \midrule[1.2pt]
\multirow{6}*{\rotatebox{90}{{\bfseries ASUMayo}}} & \multirow{3}*{\tabincell{c}{$Image$ \\ -$based$}} & $2$ & Faster R-CNN~\cite{girshick2015fast} $(ICCV'15)$ & $95.8$ & $98.8$ & $97.2$ & $78.4$ & $98.4$ & $87.3$ \\
~ & ~ & $3$ & R-FCN~\cite{dai2016r} $(NIPS'16)$ & $96.1$ & $96.4$ & $96.3$ & $80.1$ & $96.2$ & $87.4$ \\
~ & ~ & $4$ & RetinaNet~\cite{ross2017focal} $(CVPR'17)$ & $98.8$ & $84.0$ & $90.8$ & $91.8$ & $83.8$ & $87.6$ \\
~ & ~ & $6$ & FCOS~\cite{tian2019fcos} $(ICCV'19)$ & $99.5$ & $68.0$ & $80.8$ & $95.7$ & $65.4$ & $77.7$ \\
~ & ~ & $7$ & PraNet~\cite{fan2020pranet} $(MICCAI'20)$ & $98.7$ & $82.3$ & $89.8$ & $94.8$ & $82.1$ & $87.9$ \\ \cline{3-10}
~ & \multirow{3}*{\tabincell{c}{$Video$ \\ -$based$}} & $3^*$ & FGFA~\cite{zhu2017flow} $(ICCV'17)$ & $98.3_{\uparrow\bm{2.2}}$ & $91.4_{\downarrow5.0}$ & $94.8_{\downarrow1.5}$ & $88.2_{\uparrow8.1}$ & $91.1_{\downarrow5.1}$ & $89.6_{\uparrow2.2}$ \\
~ & ~ & $2^*$ & RDN~\cite{deng2019relation} $(ICCV'19)$ & $97.9_{\uparrow2.1}$ & $94.1_{\downarrow4.7}$ & $95.9_{\downarrow1.3}$ & $87.1_{\uparrow\bm{8.7}}$ & $93.7_{\downarrow4.7}$ & $90.3_{\uparrow3.0}$ \\
~ & ~ & $2^{**}$ & MEGA~\cite{chen2020memory} $(CVPR'20)$ & $96.8_{\uparrow1.0}$ & $95.9_{\downarrow2.9}$ & $96.3_{\downarrow0.9}$ & $82.6_{\uparrow4.2}$ & $94.3_{\downarrow4.1}$ & $88.1_{\uparrow0.8}$ \\
~ & ~ & $6^*$ & {\bfseries STFT $(Ours)$} & $\bm{98.9}_{\downarrow0.6}$ & $\bm{97.8}_{\uparrow\bm{29.8}}$ & $\bm{98.3}_{\uparrow\bm{17.5}}$ & $\bm{99.2}_{\uparrow3.5}$ & $\bm{97.4}_{\uparrow\bm{32.0}}$ & $\bm{98.3}_{\uparrow\bm{20.6}}$ \\
\bottomrule[2pt]
\end{tabular}
}
\end{table}

\section{Experiments}
\subsection{Datasets and Settings}
We evaluate the proposed STFT on two public video format polyp detection benchmarks.
(1) CVC-VideoClinicDB~\cite{bernal2018polyp}: 18 video sequences were split into test sets (4 videos, number of \#2, 5, 10, 18; 2484 images) and training sets (the rest 14 videos; 9470 images) following~\cite{zheng2019polyp};
(2) ASU-Mayo Clinic Colonoscopy Video~\cite{tajbakhsh2015automated}: 10 annotated videos containing polyps were split into test sets (4 videos, number of \#4, 24, 68, 70; 2098 images) and training sets (the rest 6 videos; 3304 images).
All methods in our experiments follow the same data partitioning strategy.

We use ResNet-50 \cite{he2016deep} as our backbone and FCOS \cite{tian2019fcos} as our baseline for all experiments.
STFT is trained on $4$ Tesla V100 GPUs by synchronized SGD, with one target frame and $N$ support frames holding in each GPU.
$N$ is limited by GPU memory.
We adopt a temporal dropout~\cite{zhu2017flow}, that is, randomly discard support frames in the neighborhoods $[-9,9]$ around the target frame.
We set $N=10$ in inference but $2$ in training by default.
%
%
%
%
The model will be deployed on SenseCare \cite{duan2020sensecare}.
For more training details and external experiments, please refer to \url{https://github.com/lingyunwu14/STFT}.

\begin{figure}[t]
\centering
\includegraphics[width=0.9\textwidth]{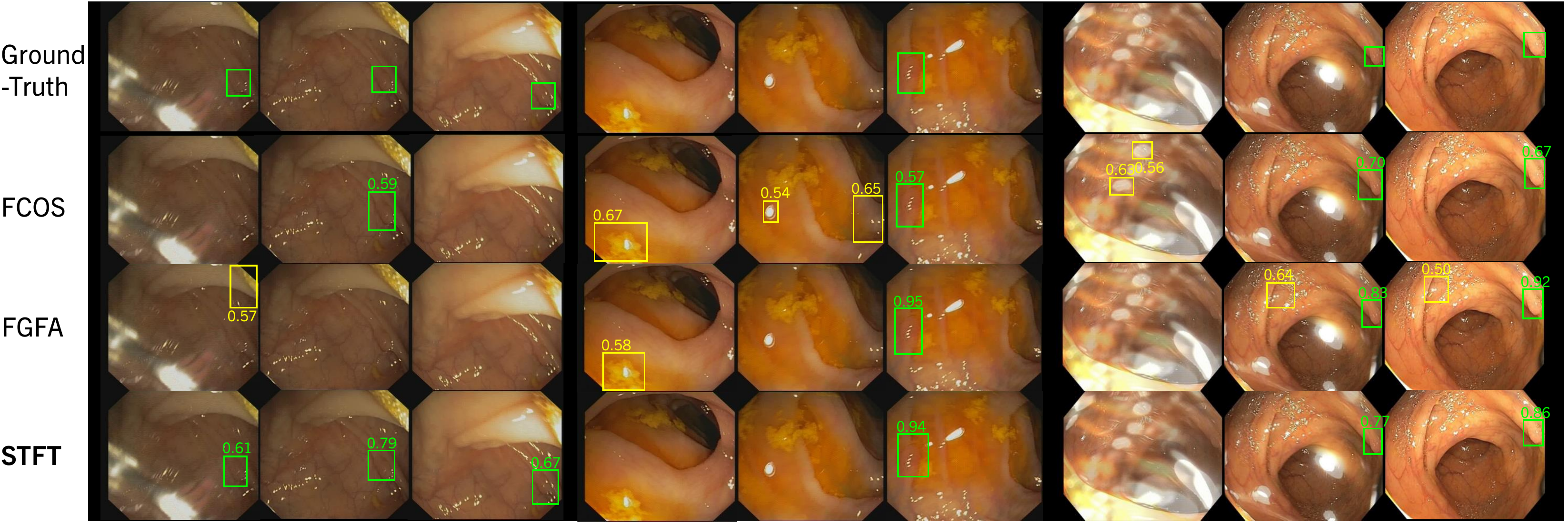}
\caption{Qualitative comparison of polyp localization. The green and yellow boxes denote correct and incorrect detections, respectively.} \label{fig_qa_image_visu}
\end{figure}

\subsection{Quantitative and Qualitative Comparison}
Table~\ref{tab_polyp} shows performance comparisons between state-of-the-art methods without any post-processing on two polyp datasets.
Following~\cite{zheng2019polyp}, precision, recall, and F1-score (the harmonic mean of precision and recall) are evaluated on two different tasks: polyp detection and polyp localization.
%
%
%
%
All compared methods are divided into two groups, image-based and video-based.
The number in front of the method represents the correspondence between video-based methods and image-based methods, such as the static baseline of $\#1^*$ is $\#1$, $\#2$ is the baseline of $\#2^*$ and $\#2^{**}$, STFT's baseline is $\#6$, etc.

Overall, STFT outperforms all SOTAs across both two datasets and two tasks, in the comprehensive metric F1-score.
On the ASUMayo, STFT performs best in all metrics.
On the CVC-Clinic, $\#1^*$ achieves higher recall, but its lower precision means a higher false-positive rate, which is not acceptable in clinical practice.
Second, STFT achieves a larger overall performance gain (F1-score: about $\geq10\%$) relative to its image-level baseline than other video-based methods.
Our baseline $\#6$ shows the lowest recall on two tasks; STFT improves it to a comparable level to the SOTAs and outperforms all methods on the ASUMayo. This suggests that STFT has a strong learning ability to effectively detect polyps.
Moreover, Fig.~\ref{fig_qa_image_visu} provides qualitative comparisons of STFT with the baseline $\#6$ and the flow-based method $\#3^*$.
STFT can precisely locate polyps in various challenging cases, such as water flow, floating content, and bubbles.

\begin{table} [t]
\caption{The effects of each module in our STFT design.}\label{tab_module_design}
\centering
\resizebox{1.0\columnwidth}{!}{
\smallskip\begin{tabular}{p{1.2cm}<{\centering}|p{3.0cm}<{\centering}p{2.2cm}<{\centering}p{2.5cm}<{\centering}p{2.4cm}<{\centering}|p{1.5cm}<{\centering}p{1.5cm}<{\centering}p{1.5cm}<{\centering}}
\toprule[1.5pt]
Methods & Temporal Aggregation? & Channel-Aware? & Spatial Adaptation? & Proposal-Guided? & Precision & Recall & F1 \\
\midrule[1.2pt]
(a) &  &  &  &  & $94.7$ & $70.4$ & $80.8$ \\
\hline
(b) & \checkmark &  &  &  & $64.4_{\downarrow30.3}$ & $77.9_{\uparrow7.5}$ & $70.5_{\downarrow10.3}$ \\
(c) & \checkmark & \checkmark &  &  & $94.6_{\downarrow0.1}$ & $82.5_{\uparrow12.1}$ & $88.2_{\uparrow7.4}$ \\
(d) & \checkmark & \checkmark & \checkmark &  & $94.3_{\downarrow0.4}$ & $83.7_{\uparrow13.3}$ & $88.7_{\uparrow7.9}$ \\
(e) & \checkmark & \checkmark & \checkmark & \checkmark & $\bm{95.0}_{\uparrow\bm{0.3}}$ & $\bm{88.0}_{\uparrow\bm{17.6}}$ & $\bm{91.4}_{\uparrow\bm{10.6}}$ \\
\bottomrule[1.5pt]
\end{tabular}
}
\end{table}

\subsection{Ablation Study}
\subsubsection{STFT Module Design}
Table~\ref{tab_module_design} compares our STFT (e) and its variants with the image-based baseline (a).
Metrics are evaluated on the polyp localization task with the CVC-Clinic dataset.
Method (b) is a naive temporal aggregation approach that directly adds adjacent features together.
The F1-score decreases to $70.5\%$.
%
Method (c) adds our channel-aware transformation into (b) for adaptive weighting.
It obtains an F1-score of $88.2\%$, $17.7\%$ higher than that of (b).
This indicates that it is critical to consider the quality weight of adjacent features.
Method (d) is a degenerated variant of (e).
It uses the original deformable convolution~\cite{dai2017deformable} to achieve spatial adaptation without our proposal-guided.
It has almost no improvement compared to (c).
(e) is the proposed STFT, which adds the proposal-guided spatial transformation module to (d).
It increases the F1-score by $10.6\%$ to $91.4\%$. The improvement for the recall is more significant ($70.4\%$ to $88.0\%$).
This proves that our proposal-guide transformation plays a key role, and STFT effectively mines useful feature representations in the neighborhood.

\begin{table}[t]
\caption{Performance and complexity comparisons in different weighting manners.}\label{tab_as_attention}
\centering
\resizebox{0.95\columnwidth}{!}{
\smallskip\begin{tabular}{p{4.6cm}|p{2.2cm}<{\centering}p{2.8cm}<{\centering}|p{1.8cm}<{\centering}p{1.8cm}<{\centering}p{1.8cm}<{\centering}}
\toprule[1.5pt]
Methods & Params & Complexity & Precision & Recall & F1 \\ 
\midrule[1.2pt]
STFT-$CosineSimilarity$~\cite{bertasius2018object,zhu2017flow} & - & $\mathcal{O}(NHW)$ & $94.2_{\downarrow0.5}$ & $79.4_{\uparrow7.0}$ & $86.2_{\uparrow5.4}$ \\ 
STFT-$PointWise$~\cite{fu2019dual} & $\alpha$ & $\mathcal{O}(NH^2W^2)$ & $93.2_{\downarrow1.5}$ & $81.7_{\uparrow11.3}$ & $87.1_{\uparrow6.3}$ \\ 
STFT-$ChannelWise$~\cite{fu2019dual} & $\beta$ & $\mathcal{O}(NC^2)$ & $\bm{95.2}_{\uparrow\bm{0.5}}$ & $80.5_{\uparrow10.1}$ & $87.3_{\uparrow6.5}$ \\ 
STFT-$\bm{ChannelAware}$ & - & $\mathcal{O}(NC)$ & $95.0_{\uparrow0.3}$ & $\bm{88.0}_{\uparrow\bm{17.6}}$ & $\bm{91.4}_{\uparrow\bm{10.6}}$ \\
\bottomrule[1.5pt]
\end{tabular}
}
\end{table}

\begin{table}[t]
\caption{Results of using different number of support frames. $^*$ indicates default setting.}\label{tab_as_support_frames}
\centering
\resizebox{0.95\columnwidth}{!}{
\smallskip\begin{tabular}{p{3.2cm}<{\centering}|p{1.0cm}<{\centering}p{1.0cm}<{\centering}p{1.0cm}<{\centering}p{1.0cm}<{\centering}p{1.0cm}<{\centering}|p{1.0cm}<{\centering}p{1.0cm}<{\centering}p{1.0cm}<{\centering}p{1.0cm}<{\centering}p{1.0cm}<{\centering}}
\toprule[1.5pt]
\# Training Frames & \multicolumn{5}{c}{$2^*$} & \multicolumn{5}{c}{$6$} \\
\# Inference Frames & $2$ & $6$ & $10^*$ & $14$ & $18$ & $2$ & $6$ & $10$ & $14$ & $18$ \\
\midrule[1.2pt]
CVC-Clinc & $91.1$ & $91.3$ & $91.4$ & $\bm{91.5}$ & $91.4$ & $90.3$ & $90.5$ & $90.7$ & $90.6$ & $90.7$ \\
ASUMayo & $98.2$ & $98.2$ & $\bm{98.3}$ & $98.3$ & $98.3$ & $95.6$ & $95.7$ & $95.8$ & $95.8$ & $95.8$ \\
\bottomrule[1.5pt]
\end{tabular}
}
\end{table}

\begin{table}[t]
\caption{Results of using different ratios of annotation frames in training.}\label{tab_as_sparse_training}
\centering
\resizebox{0.95\columnwidth}{!}{
\smallskip\begin{tabular}{p{3.2cm}<{\centering}|p{1.0cm}<{\centering}p{1.0cm}<{\centering}p{1.0cm}<{\centering}p{1.0cm}<{\centering}p{1.0cm}<{\centering}p{1.0cm}<{\centering}p{1.0cm}<{\centering}p{1.0cm}<{\centering}p{1.0cm}<{\centering}p{1.0cm}<{\centering}p{1.0cm}<{\centering}}
\toprule[1.5pt]
\# Ratios & 1 & 1/2 & 1/4 & 1/6 & 1/8 & 1/10 & 1/12 & 1/14 & 1/16 & 1/18 & 1/20 \\
\midrule[1.2pt]
Detection F1-score & $92.0$ & $92.1$ & $92.6$ & $91.9$ & $91.4$ & $\bm{92.9}$ & $91.2$ & $91.4$ & $90.2$ & $91.2$ & $90.9$ \\
Localization F1-score & $\bm{91.4}$ & $91.3$ & $91.4$ & $91.1$ & $90.8$ & $91.0$ & $90.9$ & $90.5$ & $90.3$ & $90.4$ & $90.3$ \\
\bottomrule[1.5pt]
\end{tabular}
}
\end{table}

\subsubsection{Effectiveness of Channel-Aware} \label{attention}
As noted in Table~\ref{tab_as_attention}, we use various adaptive weighting manners to replace the proposed channel-aware in temporal transformation for comparison.
In existing video detection works\cite{bertasius2018object,zhu2017flow}, calculating cosine similarity is a common weighting method.
Point-wise (in Fig.~\ref{fig_attention}(a)) and Channel-wise (in Fig.~\ref{fig_attention}(b)) are the mainstream attention mechanisms~\cite{fu2019dual}.
Compared with them, Channel-aware has the lowest computational complexity without any hyperparameters.
In addition, Channel-aware achieves the largest gain over the baseline (a) in Table~\ref{tab_module_design}.

\subsubsection{Impact of Support Frame Numbers}
We investigated the impact of different support frame numbers on STFT in Table~\ref{tab_as_support_frames}.
Under the localization F1-score metric on two datasets, training with 2 frames achieves better accuracy (6
frames reach the memory cap).
For inference, as expected, performance improves slowly as more frames are used and stabilizes.
Combining Table~\ref{tab_polyp}, STFT always achieves the highest localization F1-score and is insensitive
to support frame numbers.

\subsubsection{Learning under Sparse Annotation}
It is worth noting that we only use the ground-truth of target frames to optimize all losses of STFT.
Considering that clinical annotation is very expensive, target frames in training set are uniformly sampled to verify the learning capacity of STFT in the case of sparse labeling.
Combining Table~\ref{tab_as_sparse_training} and~\ref{tab_polyp},
STFT shows stable comprehensive performance in both detection and localization tasks.

\section{Conlusion}
We propose Spatial-Temporal Feature Transformation (STFT), an end-to-end multi-frame collaborative framework for automatically detect and localize polyp in endoscopy video.
Our method enhances adaptive spatial alignment and effective temporal aggregation of adjacent features via proposal-guided deformation and channel-aware attention.
Extensive experiments demonstrate the strong learning capacity and stability of STFT.
Without any post-processing, it outperforms all state-of-the-art methods by a large margin across both two datasets and two tasks, in the comprehensive metric F1-score.

\subsection*{Acknowledgments}	
This work is partially supported by the funding of Science and Technology Commission Shanghai Municipality No.19511121400, the General Research Fund of Hong Kong No.27208720, and the Research Donation from SenseTime Group Limited.

%
%
%
\bibliographystyle{splncs04}
\bibliography{paper906}
\end{document}